%% file: main.tex
\documentclass[sigconf]{acmart}

\usepackage{float}
\usepackage{graphics}
\usepackage{graphicx}
\usepackage{subcaption}
\usepackage{caption}
\usepackage{xspace}
\usepackage{array,multirow}
\usepackage[algo2e]{algorithm2e}
\usepackage{algorithmic}
\usepackage{algorithm}
\usepackage{amsmath}
\usepackage{tabularx}
\usepackage{adjustbox}
\newcommand\nd{\textsuperscript{nd}\xspace}

\newcommand\st{\textsuperscript{st}\xspace}

\usepackage{cuted}
\usepackage{capt-of}

\AtBeginDocument{%
  \providecommand\BibTeX{{%
    \normalfont B\kern-0.5em{\scshape i\kern-0.25em b}\kern-0.8em\TeX}}}

\copyrightyear{2021} 
\acmYear{2021} 
\setcopyright{acmcopyright}\acmConference[CHI '21]{CHI Conference on Human Factors in Computing Systems}{May 8--13, 2021}{Yokohama, Japan}
\acmBooktitle{CHI Conference on Human Factors in Computing Systems (CHI '21), May 8--13, 2021, Yokohama, Japan}
\acmPrice{15.00}
\acmDOI{10.1145/3411764.3445563}
\acmISBN{978-1-4503-8096-6/21/05}

\begin{document}

\title{DeepTake: Prediction of Driver Takeover Behavior using Multimodal Data}

\author{Erfan Pakdamanian}
\affiliation{
  \institution{School of Engineering\\ University of Virginia}
}
\email{ep2ca@virginia.edu}

\author{Shili Sheng}
\affiliation{
  \institution{School of Engineering\\ University of Virginia}
}
\email{ss7dr@virginia.edu}

\author{Sonia Baee}
\affiliation{
  \institution{School of Engineering\\ University of Virginia}
}
\email{sb5ce@virginia.edu}

\author{Seongkook Heo}
\affiliation{
  \institution{School of Engineering\\ University of Virginia}
}
\email{seongkook@virginia.edu}

\author{Sarit Kraus}
\affiliation{
  \institution{Department of Computer Science\ Bar-Ilan University}
}
\email{sarit@cs.biu.ac.il}

\author{Lu Feng}
\affiliation{
  \institution{School of Engineering\\ University of Virginia}
}
\email{lu.feng@virginia.edu}

\renewcommand{\shortauthors}{E. Pakdamanian et al.}


\begin{abstract}
\input{Sections/0-abstract}
\end{abstract}



\begin{teaserfigure}
  \includegraphics[width=\textwidth]{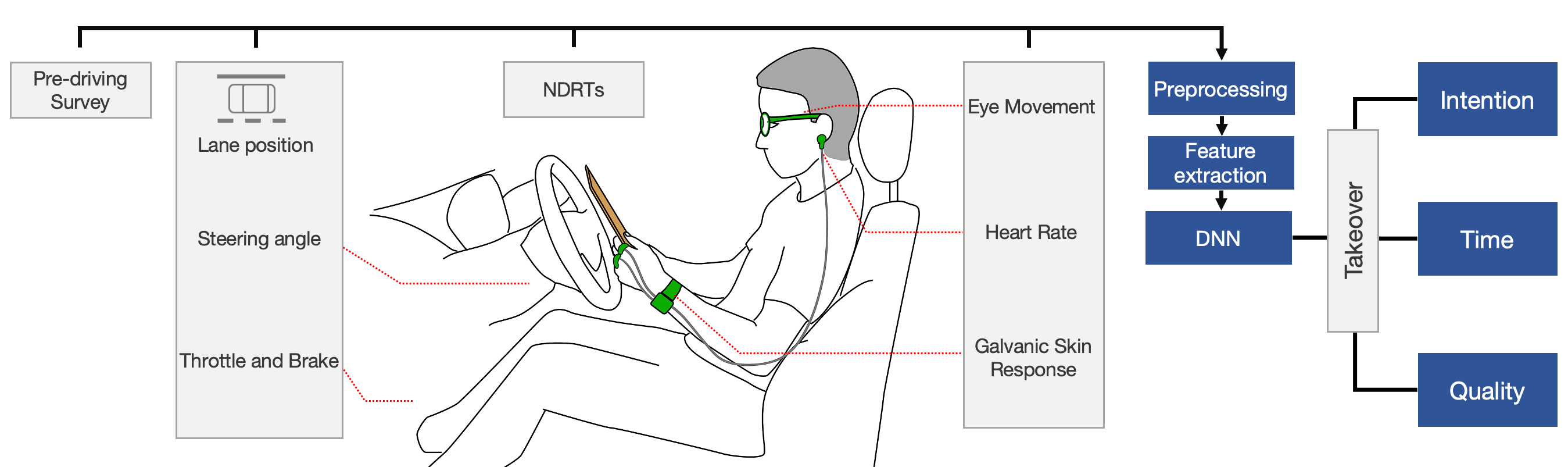}
  \caption{DeepTake uses data from multiple sources (pre-driving survey, vehicle data, non-driving related tasks~(NDRTs) information, and driver biometrics) and feeds the preprocessed extracted features into deep neural network models for the prediction of takeover intention, time and quality.}
  \label{fig:llustrative}
\end{teaserfigure}

\maketitle

\section{Introduction}
\input{Sections/1-intro}

\section{Related work}\label{sec:related}

\input{Sections/2-related-work}

\section{DeepTake: A New Approach for Takeover Behavior Prediction}
\input{Sections/3-approach}

\section{User Study} \label{sec:study}
\input{Sections/4-user-study}

\section{Performance Evaluation}\label{sec:results}
\input{Sections/5-results}

\section{Discussion} \label{sec:discussion}

\input{Sections/6-discussion}

\section{Conclusion}

\input{Sections/7-conclusion}

\section{Acknowledgment}
\input{Sections/8-Acknowledgment}

\bibliographystyle{ACM-Reference-Format}
\bibliography{References}

\end{document}

%% file: Sections/0-abstract.tex
Automated vehicles promise a future where drivers can engage in non-driving tasks without hands on the steering wheels for a prolonged period. Nevertheless, automated vehicles may still need to occasionally hand the control back to drivers due to technology limitations and legal requirements. While some systems determine the need for driver takeover using driver context and road condition to initiate a takeover request, studies show that the driver may not react to it. We present DeepTake, a novel deep neural network-based framework that predicts multiple aspects of takeover behavior to ensure that the driver is able to safely take over the control when engaged in non-driving tasks. Using features from vehicle data, driver biometrics, and subjective measurements, DeepTake predicts the driver's intention, time, and quality of takeover. We evaluate DeepTake performance using multiple evaluation metrics. Results show that DeepTake reliably predicts the takeover intention, time, and quality, with an accuracy of 96\%, 93\%, and 83\%, respectively. Results also indicate that DeepTake outperforms previous state-of-the-art methods on predicting driver takeover time and quality. Our findings have implications for the algorithm development of driver monitoring and state detection.

%% file: Sections/1-intro.tex
The rapid development of autonomous driving technologies promis-es a future where drivers can take their hands off the steering wheels and instead engage in non-driving related tasks (NDRTs) such as reading or using mobile devices. Incorporating cameras, sensors,
global positioning systems~(GPS), adaptive cruise control, light detection and ranging, and advanced driver assistance systems, automated vehicles (AVs) can navigate automatically. In Level 3 of autonomy (i.e., conditionally automated driving), as defined by the Society of Automotive Engineers (SAE international~\cite{sae2018taxonomy}), the driver does not need to continuously monitor the driving environment. Nevertheless, due to current technology limitations and legal restrictions, AVs may still need to handover the control back to drivers occasionally (e.g., under challenging driving conditions beyond the automated systems' capabilities)~\cite{mccall2019taxonomy}.
In such cases, AVs would initiate takeover requests (TORs) and alert drivers via auditory, visual, or vibrotactile modalities~\cite{naujoks2014effect,wan2018effects,pakdamanian2018effect} so that the drivers can resume manual driving in a timely manner.
However, there are challenges in making drivers safely take over control. Drivers may need a longer time to shift their attention back to driving in some situations, such as when they have been involved in NDRTs for a prolonged time~\cite{zeeb2017steering} or when they are stressed or tired~\cite{feldhutter2018wake}.
Even if TORs are initiated with enough time for a driver to react, it does not guarantee that the driver will safely take over~\cite{mcdonald2019toward}. Besides, frequent alarms could startle and increase drivers' stress levels leading to detrimental user experience in AVs~\cite{pakdamanian2020toward, korber2018have, lee2020analysis}. These challenges denote the need for AVs to constantly monitor and predict driver behavior and adapt the systems accordingly to ensure a safe takeover. 


The vast majority of prior work on driver takeover behavior has focused on the empirical analysis of high-level relationships between the factors influencing takeover time and quality (e.g., \cite{mok2017tunneled,zhang2019determinants,du2020examining,ebnali2019does}). 
More recently, the prediction of driver takeover behavior using machine learning approaches has been drawing increasing attention. 
However, only a few studies have focused on the prediction of either takeover time~\cite{lotz2018predicting,berghofer2018prediction} or takeover quality~\cite{braunagel2017ready,deo2019looking,du2020predicting,du2020predictingtake}; and their obtained accuracy results (ranging from 61\% to 79\%) are insufficient for the practical implementation of real-world applications.
This is partly due to the fact that takeover prediction involves a wide variety of factors (e.g., drivers' cognitive and physical states, vehicle states, and the contextual environment) that could influence drivers' takeover behavior~\cite{zeeb2015determines}. \par

In this paper on the other hand, we present a novel approach, named \textbf{DeepTake}, to address these challenges by providing reliable predictions of multiple aspects of takeover behavior.
\textbf{DeepTake} is a unified framework for the prediction of driver takeover behavior in three aspects:
(1) \emph{takeover intention} -- whether the driver would respond to a TOR;
(2) \emph{takeover time} -- how long it takes for the driver to resume manual driving after a TOR;
and (3) \emph{takeover quality} -- the quality of driver intervention after resuming manual control.
As illustrated in Figure~\ref{fig:llustrative}, DeepTake considers multimodal data from various sources, including driver's pre-driving survey response (e.g., gender, baseline of cognitive workload and stress levels), vehicle data (e.g., lane position, steering wheel angle, throttle/brake pedal angles), engagement in NDRTs, and driver biometrics (e.g., eye movement for detecting visual attention, heart rate and galvanic skin responses for the continuous monitoring of workload and stress levels). This data can easily be collected in AVs' driving environment. For instance, all of the driver biometrics utilized in DeepTake can be captured by wearable smartwatches and deployed eye-tracking systems. 
The multitude of sensing modalities and data sources offer complementary information for the accurate and highly reliable prediction of driver takeover behavior. 
DeepTake extracts meaningful features from the preprocessed multimodal data, and feeds them into deep neural network (DNN) models with mini-batch stochastic gradient descent.
We built and trained different DNN models (which have the same input and hidden layers, but different output layers and weights) for the prediction of takeover behavior: intention, time and quality. 
We validate DeepTake framework feasibility using data collected from a driving simulator study. Finally, we evaluate the performance of our DNN-based framework with six machine learning-based models on prediction of driver takeover behavior. The results show that DeepTake models significantly outperform six machine learning-based models in all predictions of takeover intention, time and quality.
Specifically, DeepTake achieves an accuracy of 96\% for the binary classification of takeover intention, 93\%, and 83\% accuracy for multi-class classification of takeover time and quality, respectively. These accuracy results also outperform results reported in the existing work.

The main contribution of this work is the development of DeepTake framework that predicts driver takeover intention, time and quality using vehicle data, driver biometrics and subjective measurements\footnote{DeepTake framework configurations, implementation details and code are available at \url{ https://github.com/erfpak7/DeepTake}}. The intersection between ubiquitous computing, sensing and emerging technologies offers promising avenues for DeepTake to integrate modalities into a novel human-centered framework to increase the robustness of drivers’ takeover behavior prediction.
We envision that DeepTake can be integrated into future AVs, such that the automated systems can make optimal decisions based on the predicted driver takeover behavior.
For example, if the predicted takeover time exceeds the duration that the vehicle can detect situations requiring TORs, or the predicted takeover quality is too low to respond to TORs, the automated systems can warn the driver to engage less with the NDRT. In other words, DeepTake facilitates drivers to be distracted as long as they can properly respond and safely maneuver the vehicle.
The reliable prediction of driver takeover behavior provided by DeepTake framework would not only improve the safety of AVs, but also improve drivers' user experience and productivity in AVs (e.g., drivers can focus on NDRTs without worrying about missing any TORs and potential tragic circumstances).
We believe that our work makes a step towards enabling NDRTs in automated driving, and helps HCI researchers and designers to create user interfaces and systems for AVs that adapt to the drivers’ context.

%% file: Sections/2-related-work.tex
We discuss prior work on the analysis of takeover time and quality, and position our work in the context of state-of-the-art takeover behavior prediction research. \par

\textbf{Takeover time.}
In this paper, we consider the \emph{takeover time} as the period of time from the initiation of TOR to the exact moment of the driver resuming manual control~(see Figure~\ref{fig:Scenario}), following the ISO standard definition in~\cite{iso21959}. 
Note that the same concept has also sometimes been named as takeover reaction time or response time in the literature (e.g., \cite{johns2016exploring,kim2017takeover,petermeijer2017driver,eriksson2017takeover}). The empirical literature defines a large variety of takeover time from a mean of 0.87s to brake~\cite{winter2016effects}, to an average of 19.8s to response to a countdown TOR~
\cite{politis2018evaluation} and 40s to stabilize the vehicle~\cite{merat2014transition}. This range is derived from influential factors impacting perception, cognitive processing, decision-making and resuming readiness~\cite{gold2016taking,zeeb2015determines}.
A meta-analysis of 129 studies by Zhang et al.~\cite{zhang2019determinants} found that a shorter takeover time is associated with the following factors: 
a higher urgency of the driving situation, the driver not performing a non-driving related task (NDRT) such as using a handheld device, the driver receiving an auditory or vibrotactile TOR rather than no TOR or a visual-only TOR. 
Recent studies by Mok et al.~\cite{mok2017tunneled} and Eriksson et al.~\cite{eriksson2017takeover} both confirmed that drivers occupied by NDRTs have higher responses to TORs. Similarly,~\cite{feldhutter2017duration} found a significant increase in reaction time induced by NDRTs. It is further concluded that the visual distraction causes higher reaction time when it is loaded with cognitive tasks~\cite{tang2020olfactory}. Studies have also revealed several driving environments, TOR modalities
~\cite{van2017priming, tang2020olfactory}, driving expectancy~\cite{ruscio2015does}, age~\cite{walch2017car} and gender~\cite{warshawsky2002effects} associated with takeover time. The present study extend previous findings by considering various NDRTs, gender, and objective and subjective measurements of mental workload into the DeepTake framework. \par

\textbf{Takeover quality.}
In addition to takeover time, it is essential to assess the \emph{takeover quality}, which is defined as the quality of driver intervention after resuming manual control~\cite{iso21959}. 
There are a variety of takeover quality measures, depending on different takeover situations (e.g., collision avoidance, lane-keeping), including objective measures (e.g., mean lateral position deviation, steering wheel angle deviation, metrics of distance to other vehicles or objects, minimum time to collision, frequency of emergency braking) and subjective measures (e.g., expert-based assessment, self-reported experience).
Prior work has found that takeover quality can be influenced by factors such as drivers' cognitive load~\cite{du2020evaluating, zeeb2016take}, emotions and trust~\cite{dillen2020keep,du2020examining,hergeth2017prior}, and distraction of secondary NDRTs~\cite{martelaro2019exploration,dogan2019effects}. Takeover time to an obstacle\cite{zeeb2016take} has been used widely studies as an indicator of takeover performance~\cite{eriksson2017takeover}. However, a study by Louw et al.~\cite{louw2017coming} showed that takeover time and quality appear to be independent. This lack of consensus could be due to the fact that studies apply various time budget for takeover control.

\textbf{Takeover prediction.}
While existing literature mostly focus on the empirical analysis of drivers' takeover time and quality, there are a few recent efforts on the predication of drivers' takeover behavior using machine learning~(ML) approaches. 
Lotz and Weissenberger~\cite{lotz2018predicting} applied a linear support vector machine (SVM) method to classify takeover time with four classes, using driver data collected with a remote eye-tracker and body posture camera; the results achieve an accuracy of 61\%.
Braunagel et al.~\cite{braunagel2017ready} developed an automated system that can classify the driver's takeover readiness into two levels of low and high (labeled by objective driving parameters related to the takeover quality); their best results reached an overall accuracy of 79\% based on a linear SVM classifier, using features including the traffic situation complexity, the driver's gazes on the road and NDRT involvement.
Deo and Trivedi~\cite{deo2019looking} proposed a Long Short Term Memory (LSTM) model for continuous estimation of the driver's takeover readiness index (defined by subjective ratings of human observers viewing the feed from in-vehicle vision sensors), using features representing the driver's states (e.g., gaze, hand, pose, foot activity); their best results achieve a mean absolute error (MAE) of 0.449 on a 5 point scale of the takeover readiness index. 
Du et al.~\cite{du2020predicting,du2020predictingtake} developed random forest models for classifying drivers' takeover quality into two categories of good and bad (given by subjective self-reported ratings), using drivers' physiological data and environment parameters; their best model achieves an accuracy of 70\%.

In summary, the existing works only focus on the prediction of either takeover time or takeover quality. 
By contrast, DeepTake provides a unified framework for the prediction of all three aspects of takeover behavior: intention, time and quality together. 
Furthermore, DeepTake achieves better accuracy results: 96\% for takeover intention (binary classification), 93\% for takeover time (three classes), and 83\% for takeover quality (three classes).

%% file: Sections/3-approach.tex
In this section, we present a novel deep neural network (DNN)-based approach, DeepTake, for the prediction of a driver's takeover behavior (i.e., intention, time, quality). Figure~\ref{fig:llustrative} illustrates an overvi-ew of DeepTake.
First, we collect multimodal data such as driver biometrics, pre-driving survey, types of engagement in non-driving related tasks~(NDRTs), and vehicle data. The multitude of sensing modalities and data streams offers various and complementary means to collect data that will help to obtain a more accurate and robust prediction of drivers' takeover behavior. 
Second, the collected multimodal data are preprocessed followed by segmentation and feature extraction. The extracted features are then labeled based on the belonging takeover behavior class. In our framework, we define each aspect of takeover behavior as a classification problem~(i.e., takeover intention as a binary classes whereas takeover time and quality as three multi-classes).
Finally, we build DNN-based predictive models for each aspect of takeover behavior. 
DeepTake takeover predictions can potentially enable the vehicle autonomy to adjust the timely initiation of TORs to match drivers' needs and ultimately improve safety. We describe the details of each step as follows.

\subsection{Multimodal Data Sources}\label{sec:data}

\subsubsection{Driver Biometrics}
The prevalence of wearable devices has made it easy to collect various biometrics for measuring drivers' cognitive and physiological states. 
Specifically, we consider the following three types of driver biometrics in DeepTake. \par

\textbf{Eye movement.}
Drivers are likely to engage in non-driving tasks when the vehicle is in the automated driving mode~\cite{borojeni2018reading,wintersberger2018let,pakdamanian2020toward}. 
Therefore, it is important to assess the drivers' visual attention and takeover readiness before the initiation of TORs. 
There is a proven high correlation between a driver's visual attention and eye movement~\cite{zeeb2015determines,wu2019assessing,alsaid2019moving}. DeepTake uses eye movement data (e.g., gaze position, fixation duration on areas of interest) measured by eye-tracker devices. We utilize a pair of eye-tracking glasses in our user study (see Section~\ref{sec:study}).
But the aforementioned eye movement data can be captured with any eye-tracking device. \par

\textbf{Heart rate.}
Studies have found that \emph{heart rate variability} (HRV), fluctuation of heart rate in the time intervals between the nearby beats, is a key factor associated with drivers' workload~\cite{paxion2014mental}, stre-ss~\cite{dillen2020keep}, and drowsiness~\cite{vicente2011detection}.
DeepTake uses features extracted from HRV analysis for monitoring drivers' situational awareness and readiness to respond to TORs. 
Heart rate can be measured in many different ways, such as checking the pulse or monitoring physiological signals. DeepTake employes photoplethysmographic~(PPG) signal, which can be collected continuously via PPG sensors commonly embedded in smartwatches. PPG sensors monitor heart rate by the emission of infrared light into the body and measure the reflection back to estimate the blood flow.
Unlike some heart rate monitoring devices that rely on the placement of metal electrodes on the chest, PPG sensors provide accurate heart rate measures without requiring intrusive body contact. Therefore, a PPG signal is preferred for monitoring drivers' heart rate. \par

\textbf{Galvanic skin response (GSR).}
Along with HRV, GSR has been identified as another significant indicator of drivers' stress and workload~\cite{dillen2020keep,foy2018mental,mehler2012sensitivity,radlmayr2014traffic}.
A GSR signal measures the skin conduction ability. 
Drivers' emotional arousal (e.g., stress) can trigger sweating on the hand, which can be detected through distinctive GSR patterns. 
DeepTake incorporates features extracted from the GSR signal for monitoring drivers' stress levels. 
GSR sensors are also embedded in many wearable devices, including smartwatches.

\subsubsection{Pre-Driving Survey}
In addition to the objective measurements of driver biometrics, DeepTake exploits subjective pre-driving survey responses, because drivers' prior experience and background may influence their takeover behavior~\cite{zhang2019determinants}. However, any subjective rating of factors affecting a driver's cognitive and physical ability as well as driving experience prepare a complete specification of objective metrics, potentially enhancing the distinctive attributes of an algorithm.   
DeepTake framework exerts demographic information, NASA-Task Load Index~(NASA-TLX)~\cite{hart1988development}, and the 10‐item Perceived Stress Scale~(PSS‐10)~\cite{cohen1983global} to measure drivers' perceived workload and psychological stress.
In our user study (see Section~\ref{sec:study}), we asked participants to fill in questionnaires at the beginning of each trial.

\subsubsection{Non-Driving Related Tasks (NDRTs)}
As described in Section~\ref{sec:related}, prior studies have found that engaging in NDRTs can undermine drivers' takeover performance. 
Diverse NDRTs require different levels of visual, cognitive and physical demands; thus, the influence varies when drivers are asked to interrupt the secondary task and resume manual control of the vehicle. 
DeepTake accounts for the impact of different NDRTs on the prediction of drivers' takeover behavior. 
In our user study, we considered four NDRTs in which drivers are very likely to engage in automated vehicles: (1) \textit{having a conversation with passengers}, (2) \textit{using a cellphone}, (3) \textit{reading}, and (4) \textit{solving problems} such as simple arithmetic questions~(more details in Section~\ref{sec:experimental design}). 
We chose these NDRTs because they are commonly used in driving studies~\cite{gerber2020self,dogan2019effects}, and they follow the framework of difficulty levels in the flow theory~\cite{csikszentmihalyi1990flow}. 
We further designed reading and arithmetic problem solving with two difficulty levels~(easy and medium adapted from~\cite{nourbakhsh2012using}, which reported a strong correlation between the questions and the physiological responses). 
Nevertheless, DeepTake framework can be easily adjusted to any NDRTs.

\subsubsection{Vehicle Data}
DeepTake also considers a wide range of data streams captured from the automated vehicles, including lane position, distance to hazards, angles of the steering wheel, throttle and brake pedal angles, and the vehicle velocity. Such vehicle data can help to determine the driving condition, the urgency of a takeover situation, and the impact of drivers' takeover behavior.

\subsection{Data Preparation}

\subsubsection{Feature Extraction and Multimodal Data Fusion}\label{sec:extraction}
The goal of DeepTake is to provide a procedure to reliably predict drivers' takeover behavior~(i.e., intention, time and quality) before a TOR initiation. Hence, the taken procedure for data preparation depends on the driving setting, collected data and the context. Herein, we incorporate data of drivers' objective and subjective measurements, as well as vehicle dynamic data. We initially apply data preprocessing techniques including outliers elimination, missing value imputation using mean substitutions, and smoothing to reduce artifacts presented in raw data. It is worth mentioning that we exclude any data stream providing insights about the unknown future~(e.g., type of alarm) or containing more than 50\% missing value. The preprocessed time series data are then segmented into 10-second fixed time windows \textit{prior to the occurrences of TORs}. In other words, if TOR happened at time t, we only used data captured in the fixed time window of [t-10s, t] and did not include any data later than t. We started with time window values of 2s and 18s, suggested in the literature~\cite{du2020predictingtake,braunagel2017ready,zhang2019determinants}, and experimentally settled on 10s, as real-world applications require a shorter time window with better prediction. We then aggregated the values of all multimodal data over this time interval, resulting in $256$ (max sampling rate)$\times 10sec=2560$ observations \textit{per takeover} event. However, depending on specific applications and contextual requirements, the selected time window length could vary. Subsequently, the segmented windows from modalities are processed to extract meaningful features describing the attributes impacting takeover behavior. 



\input{Tables/Features_list}

For the eye movement, we acquire interpolated features extracted from raw data through iMotion software~\cite{imotions}. The extracted eye movement attributes include gaze position, pupil diameters of each eye, time to first fixation, and fixation duration/sequence on the detected area of interest~(i.e., cellphone, tablet and monitor).

To compute the heart rate features, we first apply a min-max normalization on the raw PPG signal, and then filter the normalized PPG signal by applying a 2\nd order Butterworth high pass filter with a cut-off of 0.5Hz followed by a 1\st order Butterworth low pass filter with a cut-off frequency of 6Hz.
We use an open-source toolkit HeartPy~\cite{van2019heartpy} to filter the PPG signals and extract the following features from heart rate variability (HRV) analysis: the standard deviation of normal beats~(SDNN), root mean square of successive differences between normal heartbeats~(RMSSD), and the proportion of pairs of successive beats that differ by more than 50ms (pNN50). These metrics are to correlate with driver's cognitive workload and stress~\cite{peruzzini2019transdisciplinary}.

Furthermore, we obtain two common and important GSR features: the number and amplitude of peaks~\cite{manawadu2018tactical,nourbakhsh2012using}. A peak occurs when there is a quick burst of raised conductance
level. The peak amplitude measures how far above the baseline the peak occurred. Thus, peaks are valuable indicator of stress and mental workload.

While the variety of a driver's subjective and objective measurements along with vehicle dynamic data provide complementary information to draw better insights into drivers' takeover behavior, we need to finally fuse these multimodal data into a joint representation as input to the DNN model. Beforehand, however, we employ the Z-score normalization for most of the features except extracted PPG features to accentuate key data and binding relationships within the same range. To normalize the features associated with PPG, we use the min-max normalization, as explained above. For any remaining features still containing missing values, their missing values are imputed by using their means. Table~\ref{tbl:data sources} summarizes the list of data sources and extracted features used in DeepTake. 
Finally, the generated features from each modality concatenated to create a rich vector representing driver takeover attributes. The joint representations of all feature vectors with the provision of their associated labels are eventually fed into DNN models for training. Below, the labeling procedure of these feature vectors is explained.

\subsubsection{Data Labeling}
The target labels greatly depend on the context in which the labels are presented. Herein, we define the ground truth labeling for an attribute set denoting the feature vector. Each label indicates the classification outcome of takeover intention, time, and quality that is more representative of our user study and the three takeover behavior aspects. \par

\textbf{Takeover intention.}
DeepTake classifies a driver's takeover intention into the binary outcomes, indicating whether or not the driver would resume manual control of the vehicle. In our user study, if a participant initiated the takeover action by pressing the two buttons mounted on the steering wheel~(see Figure~\ref{fig:simulator}) upon receiving a TOR, we label the feature vector as ``TK'',  showing the takeover intention; if no takeover action was initiated between the moment of TOR initiation and the incident (e.g., obstacle avoidance), we use a ``NTK'' label displaying the absence of intention. \par

\textbf{Takeover time.}
Recall from Section~\ref{sec:related} that takeover time is defined as the time period between a TOR and the exact moment of a driver resuming manual control. Prior works have considered the starting time of manual control as the first contact with the steering wheel/pedals~\cite{zeeb2015determines} or the takeover buttons~\cite{kim2017takeover}. 
In our user study, we timed the takeover moment once a participant pressed the two takeover buttons on the steering wheel simultaneously~(see Figure~\ref{fig:simulator}).
We categorize takeover time into three classes, using threshold values consistent with the pre-defined i\textsuperscript{th} percentile of takeover time in prior driving studies~\cite{coley2009driver}.
Let $T$ denote the takeover time, thus the labels are defined as  
``low'' when $T<2.6s$, ``medium'' when $2.6s\leq T\leq6.1s$, or ``high'' when $T>6.1s$. \par

\textbf{Takeover quality.}
As we alluded to earlier in Section~\ref{sec:related}, there are a wide range of metrics~\cite{iso21959} for measuring takeover quality, depending on the needs of various takeover scenarios. 
In our user study~(see Section~\ref{sec:study}), we consider a motivating scenario where the driver needs to take over control of the vehicle and swerve away from an obstacle blocking the same lane; meanwhile, the vehicle should not deviate too much from the current lane, risking crashing into nearby traffic.
Therefore, we measure the takeover quality using the lateral deviation from the current lane, denoted by $P$. In our study, we design a 4-lane rural highway with a lane width of $3.5m$. Therefore, we label the feature vectors into three classes of takeover quality:
``low'' or staying in a lane when $P<3.5m$, ``medium'' or maneuver the obstacle but too much deviations when $7m < P\leq 10m$, or ``high'' or maneuver safely and one lane deviates when $3.5\leq P\leq7m$.

\subsection{DNN Models for Takeover Behavior Prediction}\label{sec:DNN_Model}
DeepTake utilizes a feed-forward deep neural network (DNN) with a mini-batch stochastic gradient descent.
The DNN model architecture begins with an input layer to match the input features, and each layer receives the input values from the prior layer and outputs to the next one. 
There are three hidden layers with 23, 14, and 8 ReLu units, respectively. 
The output layer can be customized for the multi-class classification of takeover intention, takeover time and takeover quality. 
For example, for the classification of takeover quality, the output layer consists of three Softmax units representing three classes~(low-, medium-, and high-) of takeover quality. 
DeepTake framework uses Softmax cross-entropy loss with an Adam optimizer with a learning rate of 0.001 to update the parameters and train the DNN models over 400 epochs. In each iteration, DeepTake randomly samples a batch of data in order to compute the gradients with a batch size of 30. Once the gradients are computed, the initiated parameters get updated. The early stopping method set to 400 epochs prevents overfitting. In addition, DeepTakes randomly divides the given labeled data into 70\% for training~(necessary for learning the weights for each node), 15\% for validation~(required to stop learning and overtraining), and 15\% for testing~(the final phase for evaluating the proposed model’s robustness to work on unseen data). Finally, in order to address imbalanced data issues where the number of observations per class is not equally distributed, DeepTake utilizes Synthetic Minority Oversampling Technique~(SMOTE)~\cite{chawla2002smote} which uses the nearest neighbor's algorithm to generate new and synthetic data.

In summary, our DeepTake framework employs different DNN models to predict takeover intention, takeover time and takeover quality. All of the DNN models in DeepTake have the same number of inputs and hidden layers, yet different output layers and associated weights.

%% file: Tables/Features_list.tex
\begin{table}[t!]
\small\addtolength{\tabcolsep}{-3pt}
  \begin{center}
    \caption{List of extracted features used in DeepTake}
    \label{tbl:data sources}
    \begin{tabular}{c| c c c}
      \toprule 
        \textbf{Data Source} & \textbf{Feature } & \textbf{Type} & \textbf{Values} \\
        \hline \hline
         \multirow{5}{*}{Eye movement} 
         & Gaze position & float & (1920$\times$1080) \\
          & Pupil size & float & (0-7) \\ 
          & Time to first fixation & int & (1-90) \\ 
          & Fixation duration & float & (100-1500ms) \\
          & Fixation sequence & int & (1-2500) \\
          \hline
          \multirow{3}{*}{Heart rate~(PPG signal)}  
          & SDNN & float & (45-75ms) \\
          & RMSSD & float & (25-43ms)\\
          & pNN50 & float & (18-28\%)\\
          \hline 
          \multirow{2}{*}{GSR signal} 
          & Number of peaks & int & (1-6)\\
          & Amplitude of peaks & float & (0.01- 1.58µs) \\
          \hline
        \multirow{3}{*}{Pre-driving survey}
          & Gender & binary & (M-W)\\
          & NASA-TLX & categorical & (1-21) \\
          & PSS-10 & categorical & (0-4)\\
           \hline
        \multirow{1}{*}{Secondary tasks}
          & NDRTs & categorical & (C,U,R,S)$^{1}$\\
          \hline
        \multirow{7}{*}{Vehicle data}        
          & Right lane distance & float & (0.73-2.4m) \\
          & Left lane distance & float & (1.02-2.8m) \\
          & Distance to hazard & float & (98-131m) \\
          & Steering wheel angle & float & (-180-114$^{\circ}$)\\
          & Throttle pedal angle & float & (15-21$^{\circ}$) \\
          & Brake pedal angle & float & (0-17$^{\circ}$) \\
          & Velocity & float & (0-55mph) \\
         
     \bottomrule 
    \end{tabular}
  \end{center}
   {\scriptsize 1:~\textit{C}; Conversation, \textit{U}; Using cellphone, \textit{R}; Reading articles on tablet,~and
   \textit{S}: Solving arithmetic questions}
\end{table}

%% file: Sections/4-user-study.tex
To test the feasibility of our proposed DeepTake framework, we conducted a user study with 20 participants featuring takeover behavior using a driving simulator\footnote{This study complies with the American Psychological Association Code of Ethics and was approved by the Institutional Review Board at University of Virginia.}. 
The following section describes the experimental setup and design of our user study as follows.

\subsection{Participants}
In this study, 20 subjects~(11~female,~9~male) aged 18-30~(mean= 23.5, SD= 3.1) were recruited. All participants were hired through the university and were required to have normal or corrected-to-normal vision, to not be susceptible to simulator sickness, and to have at least one year of driving experience to be eligible for participation in this study. Before the experiment, participants were questioned as to their age and driving experience. None of them had prior experience of interaction with AVs. They were reminded of their right to abort their trial at any point with no question asked. Three participants' data were later excluded from the analysis, due to biometric data loss and a large amount of missing values. Participants received~\$20 to compensate for the time they spent in this study.


\begin{figure}[t!]
\centering
  \includegraphics[width=0.47\textwidth]{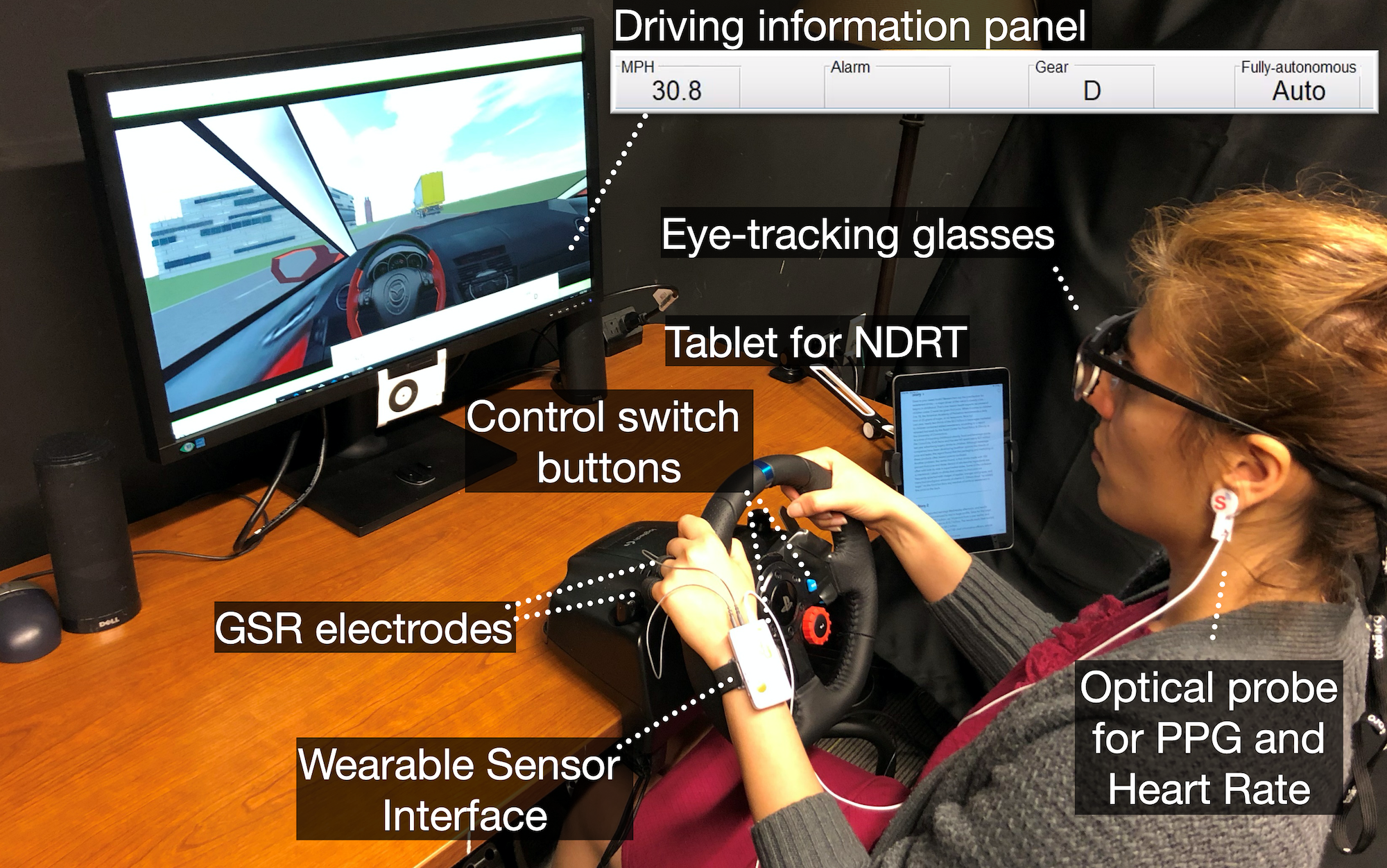}
  \caption{User study setup. This custom driving simulator consists of a 30-inch monitor, a Logitech G29 steering wheel, and 10.5-inch Apple iPad Air on which the non-driving tasks are displayed. For switching between the automated and manual control of the vehicle, the participant needs to press the two blue buttons on the steering wheel simultaneously. The participant wears a pair of eye-tracking glasses, and a wearable device with GSR and PPG sensors for the biometrics acquisition.}
  \label{fig:simulator}
\end{figure}

\subsection{Apparatus}
Figure~\ref{fig:simulator} shows our low fidelity driving simulator setup, which consists of a Logitech G29 steering wheel, accelerator, brake pedal and paddle shifters. The simulator records driver control actions and vehicle states with a sampling frequency of 20Hz and sent the captured data through a custom API using iMotions software~\cite{imotions}. The simulated driving environments along with the tasks were created using PreScan Simulation Platform. The driving environment was displayed on a 30-inch monitor. The distance between the center of the Logitech G29 steering wheel and the monitor was set at 91cm. A set of stereo speakers was used to generate the driving environment sounds along with the auditory alarm of TORs~(more details in Section~\ref{sec:experimental design}). An Apple iPad Air (10.5-inch) was positioned to the right side of the driver and steering wheel to mimic the infotainment system and displayed an article for NDRT. 

We used Tobii Pro-Glasses 2 with the sample rate of 60Hz to collect the eye movement data, and a Shimmer3+ wearable device with a sampling rate of 256Hz to measure PPG and GSR signals. To maintain consistency across all participants, we positioned the Shimmer3+ to the left of all subjects. This consistency helps reduce the motion artifact where the subjects needed to frequently interact with the tablet on the right-hand side. Although we designed our scenarios in a way to minimize the inevitable motion artifacts, we performed necessary signal processing on the PPG and GSR signals to remove potentially corrupted data, as discussed in Section~\ref{sec:data}.

\subsection{Experimental design} \label{sec:experimental design}
A within-subjects design with independent variables of stress and cognitive load manipulated by NDRTs and the TOR types was conducted with three trials in a controlled environment as shown in Figure~\ref{fig:simulator}. We designed driving scenarios in which the simulated vehicle has enough functionality similar to AVs, such that the full attention of the driver was not required at all times. \par

\textbf{Non-Driving Related Tasks.} We used four common NDRTs with various difficulty levels and cognitive demand as shown in Table~\ref{tbl:SecondaryTasks}. Participants used the tablet to read the designated articles and answer the arithmetic questions. Additionally, they were asked to use their own hand-held phones, needed for the browsing tasks. Each participant performed all NDRTs with the frequency of four times in each trial~(except for solving the arithmetic questions which occurred three times;$15\times3$ in total). The conditions and the three driving scenarios were counterbalanced among all participants to reduce order and learning effects. To have natural behavior to the greatest extent possible, participants were allowed to depart from NDRTs to resume control of the vehicle at any given time. During manual driving, participants controlled all aspects of the vehicle, including lateral and longitudinal velocity control. 
\input{Tables/NDRTs.tex}

\begin{figure*}[t!]
   \centering
   \includegraphics[width=6.5in]{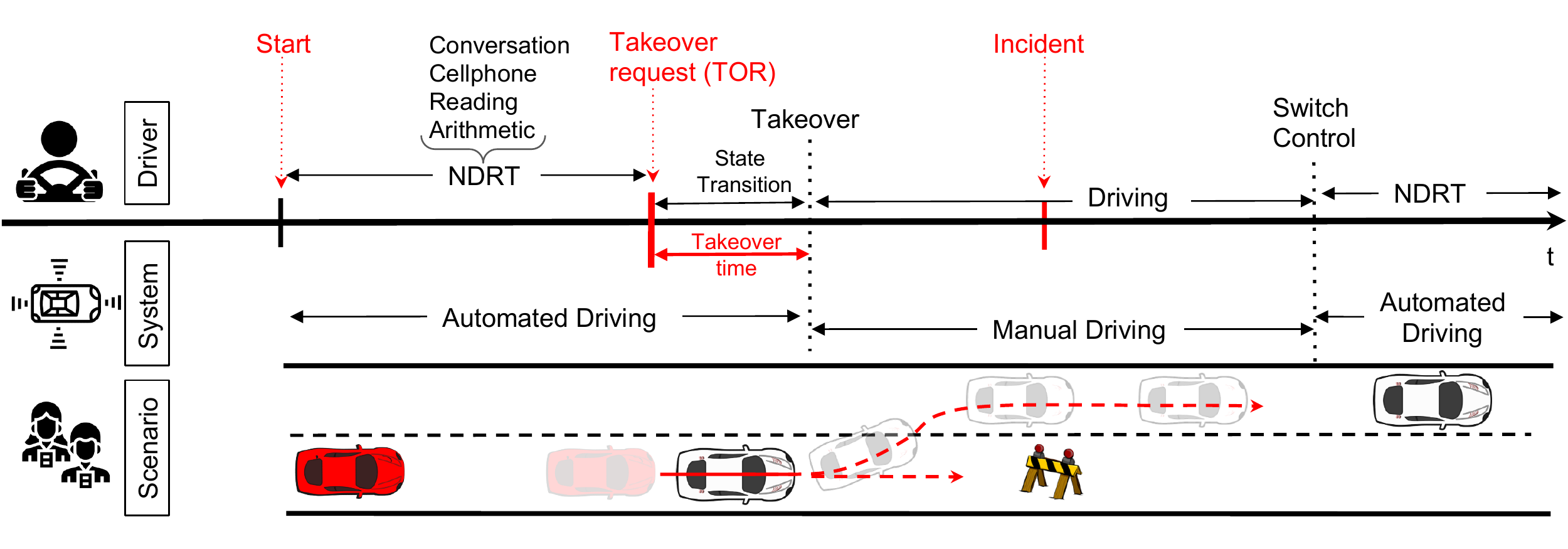}
   \caption{ A schematic view of an example of a takeover situation used in our study, consisting of: 1) takeover timeline associated with participants’ course of action; 2) system status; and 3) takeover situation. The vehicle was driven in the automated mode to the point after the TOR initiation and transitioning preparation period. The ego vehicle is shown in red and the lead car is white. When the Ego vehicle reaches its limits, the system may initiate~(true alarm) or fail~(no alarm) to initiate the TOR, and the driver takes the control back from the automated system.}
   \label{fig:Scenario}
\end{figure*} 
\textbf{Driving Scenarios.} The driving scenarios comprised a 4-lane rural highway, with various trees and houses placed alongside the roadway. 
We designed five representative situations where the AVs may need to prompt a TOR to the driver, including novel and unfamiliar incidents that appear on the same lane. Figure~\ref{fig:Scenario} shows an example of a takeover situation used in our study.
The designed unplanned takeovers let participants react more naturally to what they would normally do in AVs~\cite{mccall2019taxonomy} or as introduced by Kim and Yang~\cite{kim2017takeover}, participants' reaction times are in detectable categories. In other words, participants have no previous knowledge of incident appearance, which might happen among other incidents requiring situational awareness and decision-making.

\textbf{Takeover Requests.} In order to incorporate DeepTake in the design of adaptive in-vehicle alert systems in a way that not only monitors driver capability of takeover, but also to enhance takeover performance of automated driving, various types of TOR were required. An auditory alarm was used to inform participants about an upcoming hazard that required takeover from the automated system. The warning was a single auditory tone (350Hz, duration: 75ms) presented at the time of hazard detection~($\approx$140m or $\approx$13sec before the incidents, depending the speed of the vehicle). In a precarious world, AVs should be expected to fail to always provide correct TORs. Herein, the scenarios were constructed conservatively to include flawed TORs by which subjects would not over-trust the system's ability. In other words, the scenario demands that the participant be partially attentive and frequently perceive the environment. In order to cover the scenarios that one might encounter while driving an AV, we designed multiple critical types of TORs, including an explicit alarm (true alarm), silent failure (no alarm), and nuisance alarm (false alarm). True alarm indicates the situation in which the system correctly detects the hazard and triggers a TOR, no alarm represents the system's failure to identify the existing hazard, and false alarm presents misclassification of a non-hazardous situation as an on-road danger requiring takeover. We randomized the 15 TOR occurrences in each trial~(45 in total for each participant) with 6, 3, 6 repetitions for true alarm, no alarm, false alarm, respectively. 
In addition, we also designed an information panel where the participants could see the status of the vehicle along with the cause of TOR~(see Figure~\ref{fig:simulator}).

\subsection{Procedure} \label{sec:Procedure}
Upon arrival in the lab, participants were asked to sign a consent form and fill out a short demographic and driving history questionnaires. 
Subsequently, they were briefed on how the automated system functions, how to enable the system by simultaneously pressing two blue buttons on the steering wheel, and what they would experience during NDRTs. They were further instructed that if the system detected a situation beyond its own capabilities to handle, it would ask~(true alarm) or fail to ask~(no alarm) to take over control. Afterward, participants completed a short training drive along a highway for a minimum of 5 minutes to get familiar with the driving and assure a common level of familiarity with the setup, NDRTs, and auditory signals pitch. 

Once the subjects felt comfortable with the driving tasks and NDRTs, they proceeded to the main driving scenario. Prior to beginning the main experiment, we calibrated the eye-tracking glasses~(repeated at the beginning of each trial) and set participants up with the Shimmer3+ wearable device. Then, participants were required to complete the baseline NASA-TLX questionnaire followed by the PSS-10 questionnaire. The participants were also instructed to follow the lead car, stay on the current route, and follow traffic rules as they normally do. The participants were cautioned that they were responsible for the safety of the vehicle regardless of its mode~(manual or automated). Therefore, they were required to be attentive and to safely resume control of the vehicle in case of failures and TORs. Since the scenarios were designed to have three types of TORs, they needed to adhere to the given instruction whenever they felt the necessity. The given instruction enabled the drivers to respond meticulously whenever it was required and to reinforce the idea that they were in charge of the safe operation of the vehicle.  Due to the system's limitations, participants were told to maintain the speed within the acceptable range~($<47$mph). The experiment was conducted utilizing scenarios consisting of sunny weather conditions without considering the ambient traffic. The order of NDRT engagement was balanced for participants~(see Figure~\ref{fig:Scenario}).

The remainder of the experiment consisted of three trials, each containing 15 TORs, followed by a 5-minute break between trials. At the end of each trial, participants were requested to fill out the NASA-TLX.  After completion of the last trial, participants filled out the last NASA-TLX followed by a debrief and a \$20 compensation. The experiment took about one hour for each participant.

%% file: Tables/NDRTs.tex
\setlength{\abovecaptionskip}{-0.5pt}
\setlength{\belowcaptionskip}{-2pt}
\setlength\intextsep{1pt}
\begin{table}[t!]
\begin{center}
 \small\addtolength{\tabcolsep}{-2pt}
    
    \caption{Non-driving related tasks (NDRTs) used in our study} 
    \scalebox{0.74}{
    \label{tbl:SecondaryTasks} 
    \begin{tabular}{l|l}
        \toprule 
         \textbf{Task Type} & \textbf{Definition} \\
        \hline
        \hline
        Conversation with passenger & Interacting with the experimenter who sits close to the participants\\
        \hline
        Using cellphone & Interacting with their cellphones for texting and browsing\\
        \hline
        Reading articles & Reading three types of articles (i.e.easy,mid,hard) on the tablet \\
        \hline
        Solving questions & Answering 2-level arithmetic questions (i.e. easy and medium) \\
        \bottomrule
    \end{tabular}}
    \end{center}
\end{table}

%% file: Sections/5-results.tex
We evaluate the performance of DeepTake framework using the multimodal data collected from our user study.
We describe the baseline methods, metrics, results, and analysis as follows. 

\subsection{Baseline Methods}
Overall, we obtained about 2 million observations to train, test, and validate DeepTake with; 2560 observations per TOR $\times~15$ TORs per trial $\times~3$ trials $\times~17$ subjects. We evaluate the performance of DeepTake DNN-based models with six other ML-based predictive models, including Logistic Regression, Gradient Boosting, Random Forest, Bayesian Network, Adaptive Boosting~(Adaboost), and Regularized Greedy Forest~(RGF). Our process of choosing the ML models is an exploratory task with trials and tests of multiple off-the-shelf algorithms and choosing those that perform the best.
To evaluate the prediction performance of DeepTake framework with other ML models, we were obligated to utilize some feature importance techniques. The reasons to apply feature importance techniques for an ML algorithm are: to train the predictive model faster, reduce the complexity and increase the interpretability and accuracy of the model. In order to do so, after splitting the labeled data into training, testing, and validation sets~(see Section~\ref{sec:DNN_Model}), we employ the following feature importance methods on each training set: Absolute Shrinkage and Selection Operator~(LASSO), and random forest. LASSO helps us with not only selecting a stable subset of features that are nearly independent and relevant to the drivers' takeover behavior, but also with dimensionality reduction. The random forest method, on the other hand, ranks all of the features based on their importance levels with the drivers' takeover behavior. The overlapped features chosen by the two methods were used to train the ML-based classification models of takeover behavior.


\subsection{Metrics}
We apply 10-fold cross-validation on training data to evaluate the performance of selected features in the prediction of driver takeover intention, time and quality. Cross-validation provides an overall performance of the classification and presents how a classifier algorithm may perform once the distribution of training data gets changed in each iteration. In cross-validation, we utilize the training fold to tune model hyper-parameters (e.g., regularization strength, learning rate, and the number of estimators), which maximizes prediction performance. Therefore, we train predictive models with the best hyper-parameters. Cross-validation randomly partitions the training data into n subsets without considering the distribution of data from a subject in each set. A possible scenario is that data from one subject could be unevenly distributed in some subsets, causing overestimation of the prediction performance of a model. To avoid this situation, we check the subjects' identifiers in both the training and testing sets to ensure that they belong to just one group. We achieve this by forcing the subject to be in one group.
To determine the \textit{accuracy} of the binary classification of takeover intention performed by predictive models, accuracy was defined as
$Acc = \frac{TP+TN}{TP+TN+FP+FN}$
(TP, TN, FP, and FN represent True Positive, True Negative, False Positive, and False Negative, respectively). 
For the multi-class classification of takeover time and quality, we used the average accuracy per class.
We also used the metric of \emph{weighted F1 scores} given by 
\begin{equation}
    W F_1 = \sum_{n=1}^{l} \frac{2\times Pr_i\times Rc_i}{Pr_i+Rc_i}\times W_i,
\end{equation}
where $Pr_i = \frac{\sum_{i=1}^{l}\frac{TP_i}{TP_i+FP_i}}{l}$ is the precision, 
$Rc_i = \frac{\sum_{i=1}^{l}\frac{TP_i}{TP_i+FN_i}}{l}$ is the recall, and 
$W_i$ is the weight of the $i^{th}$ class depending on the number of positive examples in that class. It is worth mentioning that to deal with our imbalanced data, where the number of observations per class is not equally distributed, DeepTake framework along with ML-based predictive models use SMOTE to have a well-balanced distribution within class~(see Section~\ref{sec:DNN_Model}).

Given multiple classifiers, we use the \textit{Receiver Operating Characteristic}~(ROC) curve to compare the performance of DeepTake alongside other ML-based models. 
The ROC curve is a widely-accepted method that mainly shows the trade-off between TP and FP rates. A steep slope at the beginning of the curve shows a higher true positive~(correct) classification of the algorithm, whereas increasing the FP rate causes the curve to flatten. The ROC curve provides an effective way to summarize the overall performance of classification algorithms by its only metric, AUC. The AUC values provided in Figure~\ref{fig:ROC} can be interpreted as the probability of correctly classifying the driver takeover behavior into the candidate category compared to a random selection~(black line in Figure~\ref{fig:ROC}).

In addition, we use the \textit{confusion matrix} to further illustrate the summary of DeepTake's performance on the distinction of takeover intention, time, and quality per class.


\subsection{Results and Analysis}

Multiple classification algorithms were employed to compare the performance of DeepTake on obtaining a reliable discriminator of driving takeover behavior, including intention, time, and quality. As the prediction of driver takeover time and quality are contingent upon the driver's intention to take over from the autonomous systems after receiving TOR, the classification algorithms were initially carried out on this first stage of driver takeover prediction, followed by takeover time and quality. 

\vspace{1mm}

\input{Tables/Model_Comparision}
\vspace{0.1mm}

\begin{figure*}[t!]
\centering
   \includegraphics[width=\textwidth]{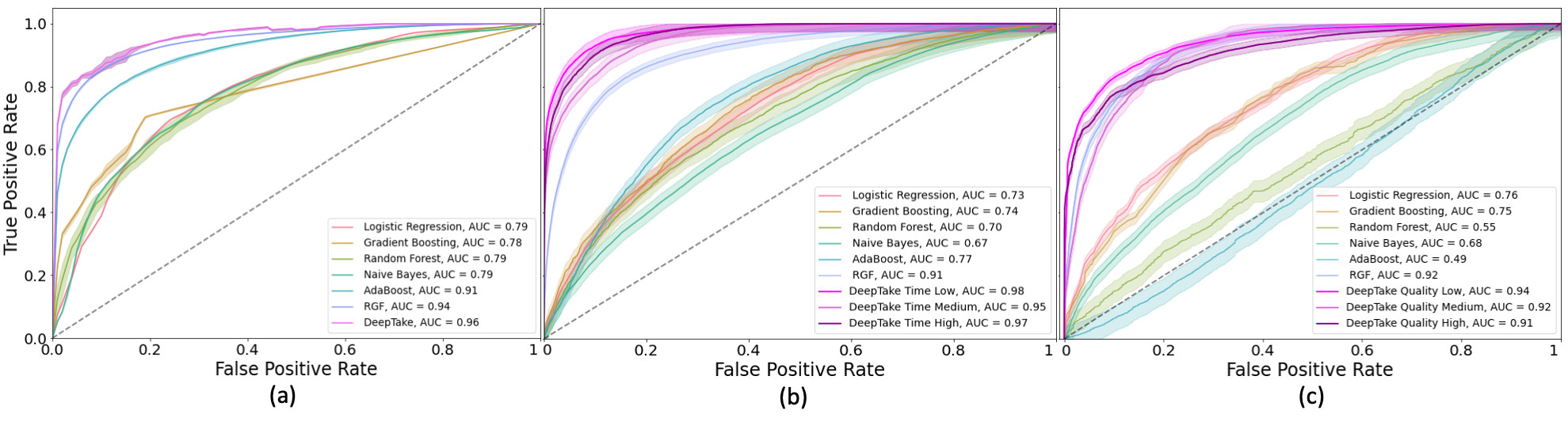}
   \caption{The ROC curve comparison of our DeepTake and six ML classification algorithms for classification of takeover behavior: (a) takeover intention, (b) takeover time, and (c) takeover quality. The ROC curve shows the average performance of each classifier and the shadowed areas represent the 95\% confidence interval. The macro AUC associated with each classifier is shown where AUC value of 0.5 refers to a chance.[Best viewed in color]}
   \label{fig:ROC}
\end{figure*} 

\begin{figure*}[t!]
\centering
  \includegraphics[width=\textwidth]{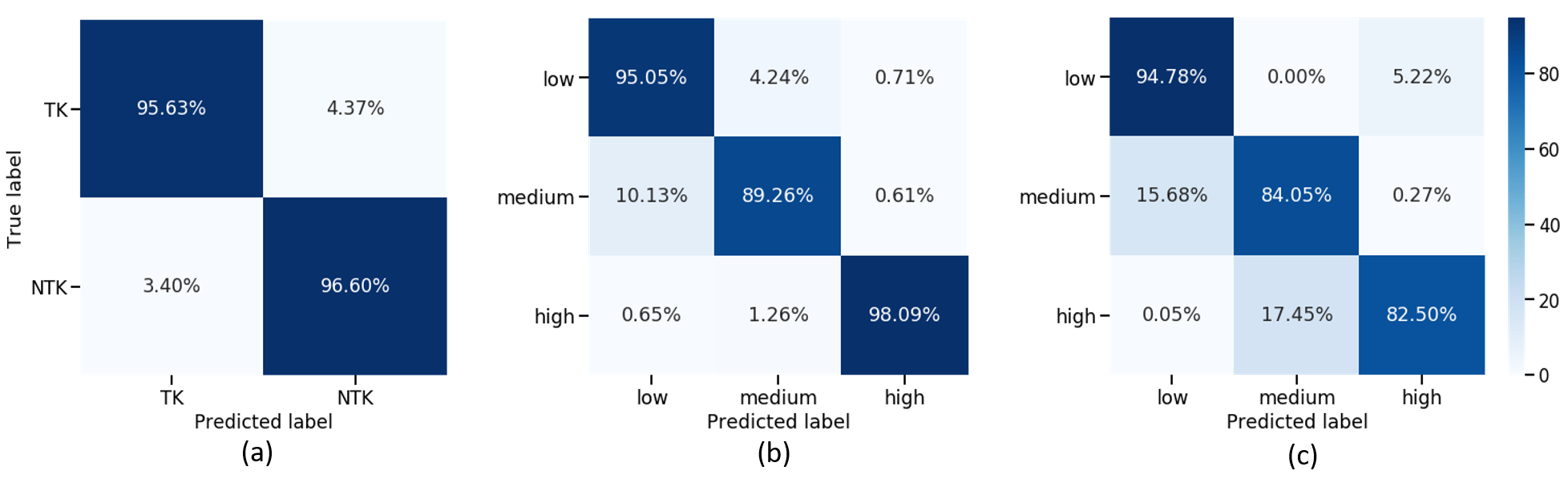}
  \caption{Confusion matrix for the prediction of takeover behavior.The results are averaged over 10 fold cross validation splits. (a) Binary class takeover intention takeover(TK) vs. Not Takeover(NTK), (b) 3-Class classification results of takeover time, (c) 3-class classification of takeover quality.}
  \label{fig:CM-All}
\end{figure*}

\textbf{Takeover intention.}
Analysis of the binary classification of drivers' takeover intention is shown in Table~\ref{tbl:results comparison}. The results show that DeepTake outperforms other ML-based models. However, among the ML-based algorithms, RGF attains the highest accuracy and weighted F1 score~(92\% and 89\%) followed by AdaBoost~(88\% and 88\%) and Logistic Regression~(77\% and 88\%). Moreover, ROC was applied in order to better evaluate each of the classifiers. Figure~\ref{fig:ROC}.a shows ROC curves and AUC values for all six ML models along with DeepTake to infer the binary classification of takeover intention. Although DeepTake shows outperformance on correctly classifying a driver's intention~(AUC=0.96) using the multimodal features, RGF shows promising performance with an AUC of 0.94. Similar to the accuracy level, AdaBoost had a slightly lower performance with an AUC= 0.91. Furthermore, we obtained the confusion matrix for takeover intention~(Figure ~\ref{fig:CM-All}.a) showing that the percentage of misclassifications is insignificant. Table~\ref{tbl:results comparison}, together with the results obtained from the AUC in Figure~\ref{fig:ROC}.a and the confusion matrix in Figure ~\ref{fig:CM-All}.a, ensure that our multimodal features with the right DNN classifier surpass the takeover intention prediction. 

\vspace{-1mm}
\begin{figure}[b!]
\centering
   \includegraphics[width=0.5\textwidth]{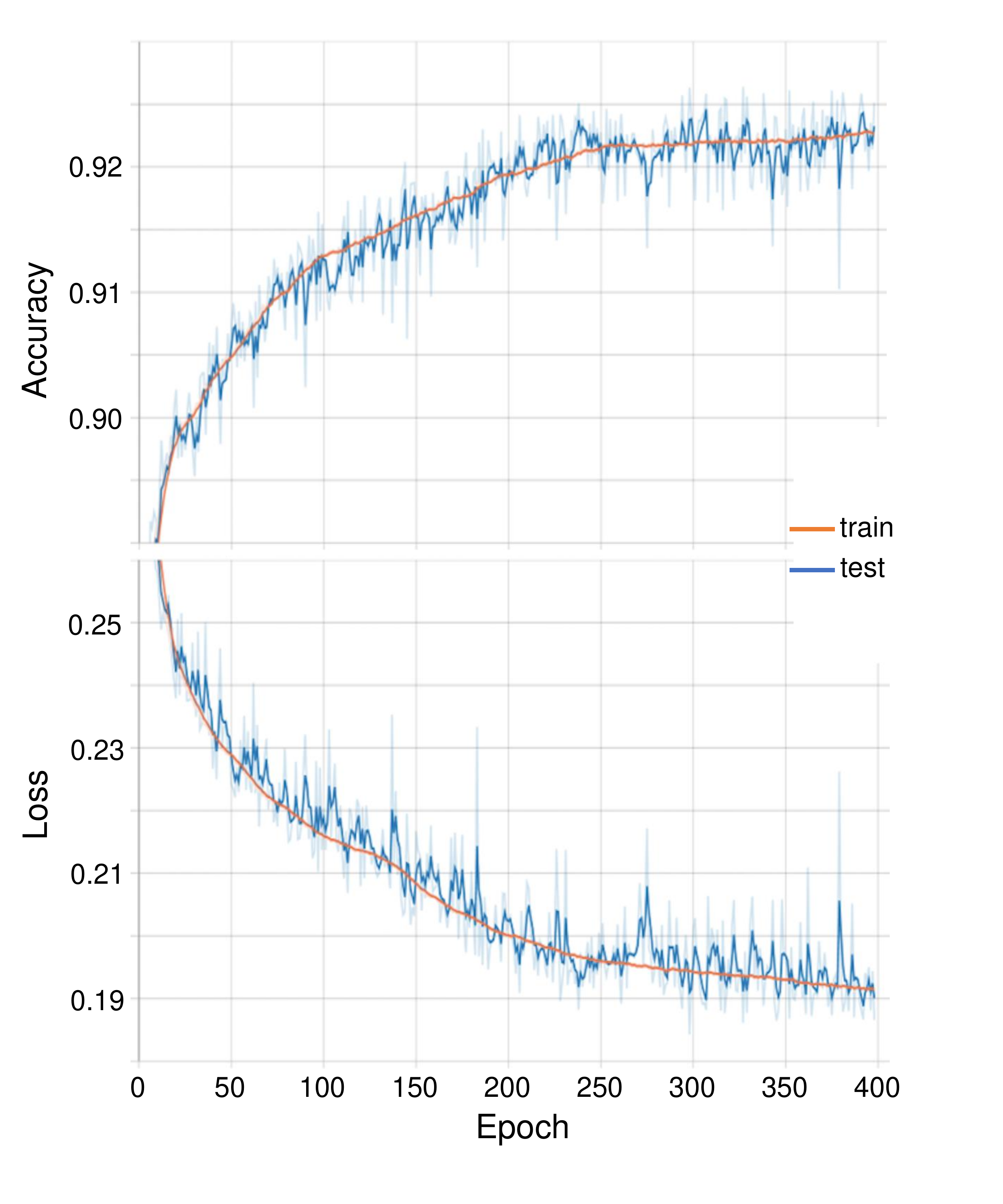}
   \caption{The top graph shows the prediction accuracy of training and test sets for 400 epochs, whereas the bottom graph indicates the loss for  DeepTake on prediction of three classes of low-, mid-, and high- takeover time.}
   \label{fig:Model_Accuracy_vs_Loss}
\end{figure} 

\textbf{Takeover time.}
DeepTake's promising performance in takeover intention estimation leads us to a more challenging multi-class prediction of driver takeover time. As some of the ML-based models attained reasonably high accuracy in the binary classification of takeover, their performances, along with our DeepTake DNN based in classifying multi-class classification of takeover time could assess the robustness of the DeepTake.

Figure\ref{fig:ROC}.b shows a comparison amongst the models explored in this paper along with DeepTake for prediction of takeover time. It displays that DeepTake produces the best overall result with an AUC value of 0.96 $\pm$ 0.02 for each takeover low-, mid-, and high- time.
We next consider the accuracy comparison of our DeepTake model with other classifier algorithms, reported in Table~\ref{tbl:results comparison}. It is evident that DeepTake outperforms all of the classic algorithms. In the three-class classification of takeover time~(low,~mid,~high), DeepTake achieves a weighted-F1 score of 0.87, thereby achieving the best performance on this task by a substantially better accuracy result of 92.8\%. Among the classifiers, RGF and AdaBoost still performed better~(73.4\% and 64.1\%). As shown in Figure~\ref{fig:Model_Accuracy_vs_Loss}, DeepTake gained a high accuracy for both the training and testing sets. However, the model did not significantly improve and stayed at around 92\% accuracy after the epoch 250.


To capture a better view of the performance of DeepTake on the prediction of each class of takeover time, we also computed the confusion matrix. Figure~\ref{fig:CM-All} displays the performance of DeepTake DNN model as the best classifier of three-class takeover time. As the diagonal values represent the percentage of elements for which the predicted label is equal to the true label, it can be seen that the misclassification in medium takeover time is the highest. Also, marginal misclassifications are found in the 2\%-5\% of the high and low takeover time classes, respectively. Overall, all three evaluation metrics of AUC, accuracy, and confusion matrix indicate that DeepTake robustness and promising performances in correctly classifying the three-class takeover time.

\textbf{Takeover quality.}
The test accuracy results of the 3-class classification of all classifiers are presented in Table~\ref{tbl:results comparison}. DeepTake achieves the highest accuracy with an average takeover quality of 83.4\%. While the value of RGF was close to DeepTake, the rest of the algorithms were not reliable enough to discriminate each class of takeover. However, we should note that RGF training time is very slow and it takes about two times longer than DeepTake to perform prediction.

In addition, Figure~\ref{fig:ROC}.c presents the multi-class classification of takeover quality. Analysis of the discriminatory properties of DeepTake achieve the highest AUC of 0.92 $\pm$ 0.01 scores among the other models for each individual class. RGF model yields an impressive average macro AUC of 0.91. Such a model indicates a high-performance achievement with informative features.

We further investigated DeepTake robustness in correctly classifying each class of takeover quality and the results achieved by the method are shown in Figure~\ref{fig:CM-All}.c. For the 3-class quality estimation, DeepTake achieved an average accuracy of 87.2\%.




%% file: Tables/Model_Comparision.tex
\begin{table}[b!]
\small\addtolength{\tabcolsep}{-3pt}
  \begin{center}
    \caption{Classification performance comparison.}
    \label{tbl:results comparison}
    \begin{tabular}{l|l| c c}
      \toprule 
        \textbf{Target value} & \textbf{Classifier } & \textbf{Accuracy} & \textbf{W-F1$^{1}$ score} \\
        \hline \hline
          \multirow{7}{*}{\shortstack{Takeover Intention}}
          & Logistic Regression & 0.77& 0.81  \\
          & Gradient Boosting & 0.76 & 0.75 \\
          & RF$^{2}$ & 0.75 & 0.72 \\
          & Naive Bayes & 0.71 & 0.66 \\
          & Ada Boost & 0.88 & 0.87 \\
          & RGF$^{3}$ & 0.92 & 0.89 \\
          &\textbf{DeepTake} & \textbf{0.96} & \textbf{0.93} \\
         
        \hline
         \multirow{7}{*}{Takeover Time}
          & Logistic Regression & 0.47& 0.45\\
          & Gradient Boosting & 0.47 & 0.46 \\
          & RF & 0.44 & 0.45 \\
          & Naive Bayes & 0.36 & 0.38 \\
          & Ada Boost & 0.64 & 0.58 \\
          & RGF & 0.73 & 0.71 \\
          & \textbf{DeepTake} &\textbf{0.93} & \textbf{0.87} \\
          \hline 
          \multirow{7}{*}{Takeover Quality}
          & Logistic Regression & 0.65 & 0.63\\
          & Gradient Boosting & 0.60 & 0.59 \\
          & RF & 0.53 & 0.52 \\
          & Naive Bayes & 0.41 & 0.39 \\
          & Ada Boost & 0.42 & 0.39 \\
          & RGF & 0.82 & 0.77 \\
          & \textbf{DeepTake} &\textbf{0.83} & \textbf{0.78}\\
     \bottomrule 
    \end{tabular}
  \end{center}
{\scriptsize 1: \textit{Weighted F1-score};
2:\textit{Random Forest};
3:\textit{Regularized Greedy Forests}} 
\end{table}

%% file: Sections/6-discussion.tex
\subsection{Summary of major findings}
In the current design of takeover requests, AVs do not account for human cognitive and physical variability, as well as their possibly frequent state changes. In addition, most previous studies emphasize the high-level relationships between certain factors and their impacts on takeover time or quality. However, a safe takeover behavior consists of a driver's willingness and readiness together. The focus of this paper is to utilize multimodal data into a robust framework to reliably predict the three main aspects of drivers' takeover behavior: takeover intention, time and quality. To the best of our knowledge, the DeepTake framework is the first method for the estimation of all three components of safe takeover behavior together within the context of AVs and it has also achieved the highest accuracy compared to previous studies predicting each aspect individually. To ensure the reliability of DeepTake's performance, we applied multiple evaluation metrics and compared the results with six well-known classifiers. Despite the promising accuracy of some of the classifiers, namely the RGF classifier, the accuracy of DeepTake surpassed in its prediction of takeover behavior. In general, our model performed better in classifying driver takeover intention, time and quality with an average accuracy of 96\%, 93\%, and 83\%, respectively.

\begin{figure}[t!]
\centering
  \includegraphics[width=0.48\textwidth]{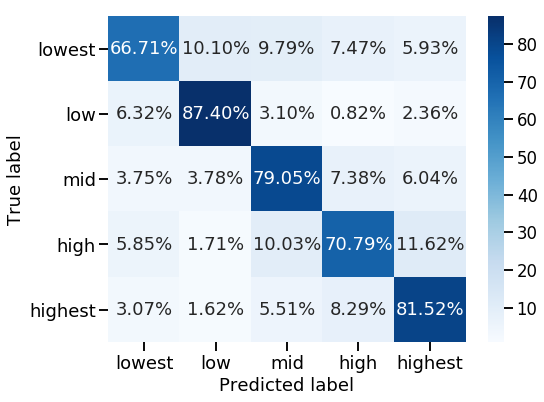}
  \caption{Confusion matrix for the prediction of five classes of driver takeover time.}
  \label{fig:CM-FiveClass}
\end{figure}

\begin{figure*}[t!]
\centering
  \includegraphics[width=7.5in]{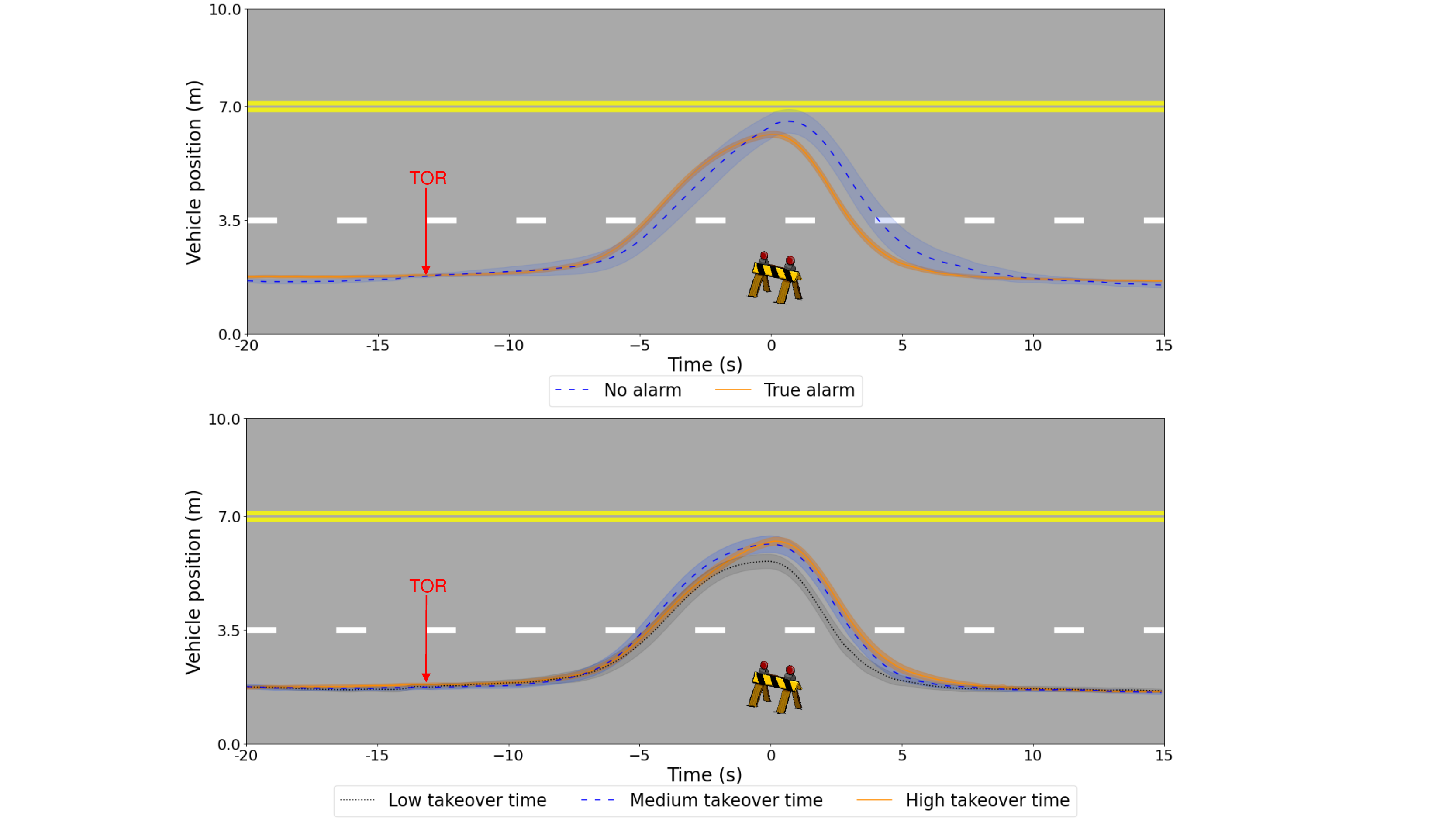}
  \vspace{1mm}
  \caption{Average trajectories when drivers took over control from automated system after receiving TORs. Top graph shows the lateral position of the vehicle with respect to no alarm~(silent failure) and true alarm~(explicit alarm). Bottom graph shows the lateral position of the vehicle for three categories of takeover time (low, mid, and high). The light shaded area representing standard deviation at each time point.}
  \label{fig:Trajectory_time} 
\end{figure*}

In order to further assess the robustness of DeepTake, we increase the number of classes to the more challenging five-class classification of takeover time where the classes defined as ``lowest'' when $T<1.5s$, ``low'' when $1.5s\leq T<2.6s$, ``medium'' when $2.6s\leq T<4.7s$, ``high'' when $4.7s\leq T\leq6.1s$, or ``highest'' when $T>6.1s$. Figure~\ref{fig:CM-FiveClass} represents the performance of DeepTake on classifying the five-class takeover time. Although DeepTake was not as distinctive in five-class classification as in the three-class, it still achieved promising results. Lowest, high, and medium takeover times are the top three pairs that were the most frequently misclassified by the  DNN model. The reason might be that the selected features do not have the required distinctive characteristics to perfectly divide the low and medium takeover time. In each class, it could still distinguish between five other classes with an average accuracy of 77\%. With a future larger amount of data collection satisfying each class need,  DeepTake could further improve its distinctive aspect of each feature for more precise classification.

\subsection{Descriptive analysis of takeover time and quality}
Although DeepTake takes advantage of a DNN-based model integrated into its framework, understanding the reasons behind its predictions is still a black-box and a challenging problem which will be tackled in our future works. However, to comprehend the effects of multimodal variables on takeover time and quality, a repeated measure Generalized Linear Mixed~(GLM) model with a significance level of $\alpha$ = 0.05 to assess the correlation of suboptimal features was used to predict takeover time and quality. The analysis of the results shows the significant main effect of NDRTs on takeover time and quality~($F_{3,28}=13.58$, $p<0.001$) followed by fixation sequence~($F_{1,28}=35.87$, $p<0.001$) and vehicle velocity~($F_{1,28}=13.06$, $p<0.001$). Post-hoc tests using Bonferroni demonstrated a higher impact of interaction with the tablet and reading articles~($p<0.001$) as opposed to a conversation with passengers. This result could be based on the amount of time spent and the level of cognitive load on the takeover task. This finding is aligned with the previous results of~\cite{feldhutter2017duration,eriksson2017takeover}. Additionally, there was no significant effect of brake and throttle pedal angle on the takeover time($F_{1,28}=3.05$, $p=0.085$) and quality~($F_{1,28}=1.27$ $p=0.256$). This could be because our scenarios did not take place on crowded roads and participants were not forced to adopt a specific behavior after the TOR. Therefore, they could maneuver the vehicle without significant adjustment to either pedal.


On the other hand, takeover quality tied into drivers' lane keeping control and was impacted by the alarm type and the category of takeover time shown in Figure~\ref{fig:Trajectory_time}. Although we did not consider the type of alarm and category of takeover time for prediction of takeover behavior as they could simply manipulate DeepTake outcomes by providing insights about the future, it is worth additional investigation of their impacts on the takeover quality. Since participants' takeover times and the speed of the vehicle on the manual driving were different, Figure~~\ref{fig:Trajectory_time} shows the average time of TOR. The top graph in Figure~\ref{fig:Trajectory_time} depicts the average lateral position of the vehicle with respect to no alarm and true alarm. These two types of the alarm were considered due to the necessity of taking over. Under the impact of the true alarm, the vehicle deviates less than when there is no alarm, yet not significantly~($F_{2,28}=7.07$, $p=0.78$). Moreover, the drivers performed more abrupt steering wheel maneuvers to change lanes on true alarm. Similarly, the bottom graph in Figure \ref{fig:Trajectory_time} shows the lateral position with respect to different takeover times~(low, mid, and high). It can be seen that the longer the takeover time is, the farther the vehicle deviates from the departure lane. Differences in takeover time were also analyzed to investigate the takeover quality. The main effect of the type of takeover time was not significant~($F_{2,19}=0.44$). Although prior research has revealed various timing efforts to fully stabilize the vehicle~\cite{merat2014transition}, our observations are comparable to~\cite{naujoks2019noncritical} and \cite{bueno2016different}.

\subsection{Implications on the design of future interactive systems}

We believe that our human-centered framework makes a step towards enabling a longer interaction with NDRTs for automated driving. DeepTake helps the system to constantly monitor and predict the driver's mental and physical status by which the automated system can make optimal decisions and improve the safety and user experience in AVs. Specifically, by integrating the DeepTake framework into the monitoring systems of AVs, the automated system infers when the driver has the intention to takeover through multiple sensor streams. Once the system confirms a strong possibility of takeover intention, it can adapt its driving behavior to match the driver's needs for acceptable and safe takeover time and quality. Therefore, a receiver of TOR can be ascertained as having the capability to take over properly, otherwise, the system would have allowed the continued engagement in NDRT or warned about it. Thus, integration of DeepTake into the future design of AVs facilitates the human and system interaction to be more natural, efficient and safe. Since DeepTake should be used in safety-critical applications, we further validated it to ensure that it meets important safety requirements~\cite{grese2021formal}. We analyzed DeepTake sensitivity and robustness with several techniques using the Marabou verification tool. The sensitivity analysis provides insight into the importance of input features, in addition to providing formal guarantees with respect to the regions in the input space where the DeepTake behaves as expected. \par
DeepTake framework provides a promising new direction for modeling driver takeover behavior to lessen the effect of the general and fixed design of TORs which generally considers homogeneous takeover time for all drivers. This is grounded in the design of higher user acceptance of AVs and dynamic feedback~\cite{seppelt2019keeping,ekman2017creating}. The information obtained by DeepTake can be conveyed to passengers as well as other vehicles letting their movement decisions have a higher degree of situational awareness.
We envision that DeepTake would help HCI researchers and designers to create user interfaces and systems for AVs that adapt to the drivers’ state.

\subsection{Limitations and future work}
The following limitations should be taken into consideration for future research and development of DeepTake. \\
First, it is acknowledged that the DeepTake dataset is vulnerable to the low fidelity driving simulator used for data collection. It is possible that the takeover behavior of subjects were influenced by the simplicity of driving setup and activities. To apply DeepTake on the road, we will need more emphasis on various user's activities and safety, and exclude subjective surveys causing biases. Second, while we increased the number of classes, future development of DeepTake should predict takeover time numerically. For this purpose, a larger dataset will be needed which accounts for a high variation of individual takeover time and probabilistic nature of DNNs by which the DeepTake framework can still learn and reliably predicts takeover time. 

Third, although we tried to avoid overfitting, it is possible that DeepTake emphasized more on few features that frequently appeared in TORs, and the performance may not be the same if more scenarios are being tested. Thus, DeepTake decision boundaries need to be experimented with different adversarial training techniques.  
Forth, DeepTake lacks using real-world data which often significantly different and could potentially impact the results of DeepTake framework. Testing the framework on real-world data helps users to gain confidence in DeepTake's performance. DeepTake was developed and assessed offline using a driving simulator in a controlled environment. Future work should explore the deployment of DeepTake online and in the wild for real-world applications in future AVs. We plan to integrate the DeepTake and its verification results~\cite{grese2021formal} into the safety controller, which will be then evaluated using the on-road vehicle. In our
future work we also plan to try to reduce the number of features in the model by using the results from the sensitivity analysis along with feature importance analysis techniques~(i.e. LIME and SHAP) to discover features that may be able to be dropped from the model. 


%% file: Sections/7-conclusion.tex
In this work, we present DeepTake, a novel method that predicts driver takeover intention, time and quality using data obtained from the vehicle, wearable sensors, and a self-administered survey taken before driving. By using DNN-based models, DeepTake enables prediction of driver takeover intention, time and quality, all of which are crucial in ensuring the safe takeover of an automated vehicle. Our evaluation showed that DeepTake outperforms the best accuracy results of prior work on takeover prediction with an accuracy of 96 \%, 93 \%, and 83\% for the multi-class classification of takeover intention, time and quality, respectively. 
As prior studies demonstrated, alarming drivers when the system detects a situation requiring takeover does not guarantee safe driver takeover behavior ~\cite{korber2018have, lee2020analysis, mcdonald2019toward}. We believe that accurate takeover prediction afforded by DeepTake would allow drivers to work on non-driving related tasks while ensuring that they safely take over the control when needed. DeepTake opens up new perspectives for HCI researchers and designers to create user interfaces and systems for AVs that adapt to the drivers’ context.


%% file: Sections/8-Acknowledgment.tex
We would like to thank Prof. Corina P\u{a}s\u{a}reanu from Carnegie Mellon University and Prof. Radu Calinescu from University of York for their valuable inputs, and John Grese for his help in evaluating DeepTake with a high number of epochs. This work was supported in part by National Science Foundation CCF-1942836 grant, Assuring Autonomy International Programme, and Toyota InfoTech Labs.